\documentclass[conference]{IEEEtran}
\IEEEoverridecommandlockouts
\usepackage{amsfonts}
\usepackage{algorithm}
\usepackage{algpseudocode}
\usepackage{array}
\usepackage[caption=false,font=normalsize,labelfont=sf,textfont=sf]{subfig}
\usepackage{textcomp}
\usepackage{stfloats}
\usepackage{url}
\usepackage{amsmath}
\usepackage{bm}
\usepackage{hyperref}
\usepackage{verbatim}
\usepackage{graphicx}
\usepackage{cite}
\usepackage{svg}
\usepackage{multirow}

\usepackage{todonotes}
\usepackage{setspace}
\usepackage{tablefootnote}
\usepackage{threeparttable}
\usepackage{siunitx}
\usepackage{multirow}
\usepackage[inline]{enumitem}

\newcommand{\ourmethod}{our method\ }

\newcommand{\email}[1]{\href{mailto:#1}{\nolinkurl{#1}}}

\def\BibTeX{{\rm B\kern-.05em{\sc i\kern-.025em b}\kern-.08em
    T\kern-.1667em\lower.7ex\hbox{E}\kern-.125emX}}
    
\begin{document}

\title{\huge Tiny Deep Ensemble: Uncertainty Estimation in Edge AI Accelerators via Ensembling Normalization Layers with Shared Weights}

\author{
Soyed Tuhin Ahmed, Michael Hefenbrock, Mehdi B. Tahoori\\
Karlsruhe Institute of Technology, Revoai GmbH, Germany\\soyed.ahmed@kit.edu
\vspace{-.4em}
}

\maketitle

\begin{abstract}

The applications of artificial intelligence (AI) are rapidly evolving, and they are also commonly used in safety-critical domains, such as autonomous driving and medical diagnosis, where functional safety is paramount.
In AI-driven systems, uncertainty estimation allows the user to avoid overconfidence predictions and achieve functional safety.
Therefore, the robustness and reliability of model predictions can be improved. However, conventional uncertainty estimation methods, such as the deep ensemble method, impose high computation and accordingly hardware (latency and energy) overhead because they require the storage and processing of multiple models. Alternatively, Monte Carlo dropout (MC-dropout) methods, although having low memory overhead, necessitate numerous ($\sim 100$) forward passes, leading to high computational overhead and latency. Thus, these approaches are not suitable for battery-powered edge devices with limited computing and memory resources. In this paper, we propose the Tiny-Deep Ensemble approach, a low-cost approach for uncertainty estimation on edge devices. In our approach, only normalization layers are ensembled $M$ times, with all ensemble members sharing common weights and biases, leading to a significant decrease in storage requirements and latency. Moreover, our approach requires only one forward pass in a hardware architecture that allows batch processing for inference and uncertainty estimation. Furthermore, it has approximately the same memory overhead compared to a single model. Therefore, latency and memory overhead are reduced by a factor of up to $\sim M\times$.
Nevertheless, our method does not compromise accuracy, with an increase in inference accuracy of up to $\sim 1\%$ and a reduction in RMSE of $17.17\%$ in various benchmark datasets, tasks, and state-of-the-art architectures.

\end{abstract}

\addtolength\abovedisplayskip{-0.6em}%
\addtolength\belowdisplayskip{-0.6em}%
\setlength{\textfloatsep}{-1pt}
\captionsetup{belowskip=1pt,aboveskip=1pt}

\begin{IEEEkeywords}
Deep Ensemble, BatchEnsemble, TinyML, Uncertainty Estimation, MC-Dropout
\vspace{-.4em}
\end{IEEEkeywords}

\section{Introduction}\label{sec:intro}

Recent advances in deep learning models, such as neural networks (NNs), have shown superior performance in various domains~\cite{lecun2015deep}. Consequently, they are widely adopted in different sectors, including critical ones such as automotive, health care, and industrial control. However, training and inference of modern NN models require a tremendous amount of computational power and memory. Therefore, they are suitable for the cloud computing paradigm due to their “unlimited” storage capacity and computing resources~\cite{carroll2011secure}, but they are challenging for edge AI accelerators. Edge AI acceleration provides privacy and real-time processing, but they have limited computational and memory resources. 
Numerous industries may expect significant transformations due to AI-powered edge computing~\cite{loven2019edgeai} and their the market is estimated to be worth $\$3.5$ billion by 2027~\cite{hu2021rim}.

Numerous contributions have been made in the field of TinyML~\cite{tinyml_survey}, where the emphasis is on the running of NNs on hardware with extremely low power, memory, and computational resources while still maintaining reasonable accuracy. Nevertheless, research on predictive uncertainty estimation with tiny NN models is lacking. 

Uncertainty estimation in prediction is crucial in safety-critical applications where NNs operate on real-time data, e.g., from sensory inputs. During NN deployment, the underlying data distribution may shift or the data may become corrupted due to sensor noise~\cite{hendrycks2019benchmarking}. To address this, predictive uncertainty can supplement model predictions and enable informed decision-making. As a result, unreliable predictions can be prevented from reaching the end user and reviewed by a human expert.


\begin{figure}
    \centering
    \includegraphics[width=0.7\linewidth]{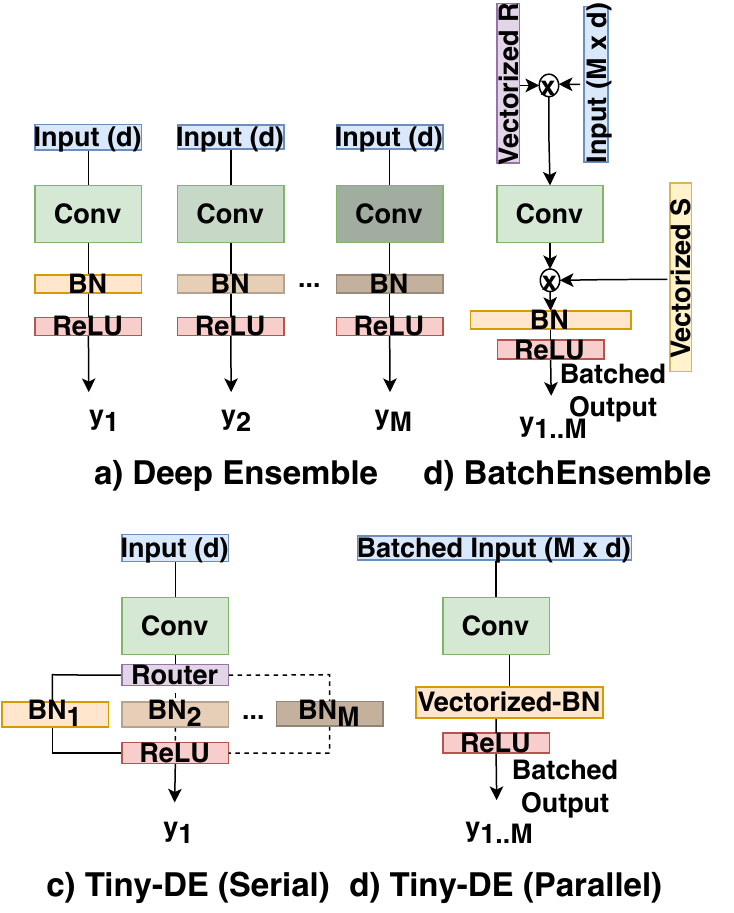}
    \vspace{-0.2em}
    \caption{a) Deep Ensemble~\cite{deep_ensemble} with $M$ ensemble members
    , b) BatchEnsemble~\cite{batchensemble}, proposed Tiny-DE model with $M$ normalization layers with \emph{a single shared convolutional layer}  in c) serial mode, and d) parallel mode. 
    }
    \label{fig:tiny_de_vs_related}
\end{figure}

Among the numerous uncertainty estimation methods~\cite{abdar2021review}, the Deep Ensemble~\cite{deep_ensemble} is considered a “gold standard” for uncertainty estimation~\cite{wilson2020bayesian}. In the Deep Ensemble, $M$ ensemble members $1, \cdots, M$ are trained independently and stored in hardware. During inference in edge AI accelerators, the input is processed by each model in $M$ forward passes (see Fig.~\ref{fig:tiny_de_vs_related} (a)). Subsequently,
the outputs of all models are combined to obtain the predictive distribution. Therefore, the cost in terms of latency, power, and memory for training, storage, and processing of $M$ ensemble members is challenging for edge AI accelerators.

\begin{figure}
    \centering
    \includegraphics[width=0.9\linewidth]{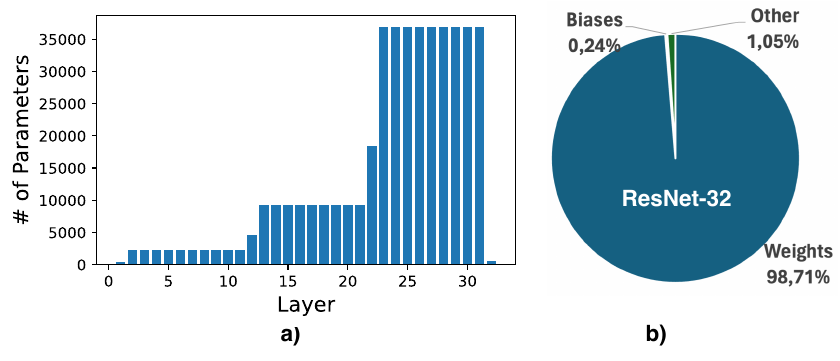}
    \vspace{-0.7em}
    \caption{a) Number of parameters in each layer and b) Share of parameter groups with respect to the total number of parameters in ResNet-32.
    }
    \label{fig:param_pie}
\end{figure}

To reduce computational and memory overhead in ensemble methods, several studies exist. 
Monte Carlo dropout (MC-dropout) can be interpreted as "implicit" ensembles that can create an exponential number
of weight-sharing sub-networks for uncertainty estimates~\cite{gal2016dropout}. 
Although MC-dropout requires training and storage of a single model, inference involves $M$ forward passes through a dropout-enabled network. Here, $M$ varies with tasks, and the topology can be as large as $94$ even on a small (six-layer) fully convolutional network~\cite{mobiny2021dropconnect}. Furthermore, MC-dropout has sampling latency and chip area overhead for the dropout module implementation~\cite{ahmed2023spindrop}. To reduce inference latency, the work~\cite{rock2021efficient} proposed to ensemble only deeper convolutional layers while the shared backbone is computed only once and cached. However, in convolutional NNs (CNNs), deeper convolutional layers have significantly larger parameter counts than other layers, as shown in Fig~\ref{fig:param_pie} (a). Also, this approach only works if dropout is applied
only to deeper convolutional layers rather than to all layers. Another domain-specific group of work, binarized the MC-Dropout to reduce memory overhead by $32\times$, improve latency by accelerating them in Spintronics-based computation-in-memory (CIM) architecture, and reduce sampling latency by reducing the total number of Dropout modules~\cite{ahmed2023spindrop, tuhin2022binary, ahmed2023scale, ahmed2023spatial}. In contrast, the BatchEnsemble~\cite{batchensemble} approach also shares weights but introduces two sets of $M$ rank-1 matrices to generate $M$ ensemble members. Their approach is not scalable to the AI accelerator architecture that does not allow batch processing. Additionally, it introduces additional computation at the input and output of a layer, as shown in Fig.~\ref{fig:tiny_de_vs_related} (b).


We observed that parameters other than weights and biases in a NN consume only $\sim 1\%$ of all parameters, as shown in Fig.~\ref{fig:param_pie} (b). Therefore, we propose to ensemble only normalization layers with shared weight and biases. 
The normalization layer is commonly used in NNs, as it speeds up the training and improves performance~\cite{ioffe2015batch}. Our approach is scalable 1) in any AI accelerator architecture, 2) in any NN topologies, such as CNN and recurrent neural network (RNN), 3) in tasks, and 4) in datasets. Furthermore, our approach is parallelizable during training and inference within an AI accelerator architecture. Consequently, all ensemble members can be updated concurrently for a given mini-batch, and inference requires a single forward pass, allowing for \emph{single-shot training and inference}.


Our contributions can be summarized as follows:
\begin{itemize}
    \item Ensembling normalization layers with shared weights and biases for low-cost uncertainty estimation, tailored for edge AI accelerators.
    \item Tiny-DE network topology that is scalable to existing NN topologies, AI accelerator architectures, and NN tasks.
    \item Single-Shot training and inference in hardware architecture that allows batch processing. 
    \item EnsembleNorm layer for normalizing all ensemble members in a single shot.  
    \item Substantial reduction in computational and storage requirements without sacrificing accuracy and quality of uncertainty estimates, as evidenced by extensive empirical evaluation.
\end{itemize}

The rest of the paper is organized as follows: Section~\ref{sec:related_work} reviews the related work, Section~\ref{sec:methodology} details the proposed methodology, Section~\ref{sec:experiments} provides experimental results, and Section~\ref{sec:conclusion} concludes the paper.
\section{Preliminary}\label{sec:related_work}

\subsection{Uncertainty In Deep Learning}

In deep learning, uncertainty estimation is crucial for evaluating the reliability and robustness of model predictions. It offers vital information about confidence in these predictions. This is especially crucial in supporting decision-making in safety-critical applications, such as autonomous driving and automatic medical diagnostics.

Deep learning models are usually deployed in dynamic and uncertain environments where the distribution of inference data can change over time. Therefore, the model can receive input data that is unseen during training and its distribution is completely different. For example, a model trained on the MNIST (handwritten digit recognition) dataset can receive corrupted data during inference due to sensor noise or domain shift. Such data are referred to as out-of-distribution (OoD) data or sometimes called out-of-training-distribution data points~\cite{guo2017calibration, jospin2022hands}. Uncertainty in prediction arises primarily due to the tendency of the model to give overconfident predictions for unknown data. For example, in a classification task, the model will predict that the unseen OoD data belong to one of the classes with close to $100\%$ confidence~\cite{ahmed2023spindrop}. In such scenarios, quantifying the uncertainty in the prediction allows the user to make informed decisions and avoid catastrophic failures. 

A reliable uncertainty estimation method should demonstrate low uncertainty in data similar to what it has been trained on, in distribution (ID) data, and 
high uncertainty on unseen or OoD data. In a fine-grained method, an incorrect prediction should show high uncertainty and a correct prediction should show low uncertainty. 

Note that there is a difference between generalizing on the same data, i.e., inference accuracy, and OoD data. Inference accuracy refers to prediction accuracy with data that have the same distribution as training data but are unseen during training, e.g., validation data. An ideal uncertainty estimation method, during inference, is expected to generalize well on the same data distribution and provide interpretable uncertainty estimates on OoD data.

\subsection{Normalization Approaches}

In modern-deep learning topologies, normalization layers are essential to improve training stability, speed, convergence, and performance~\cite{ioffe2015batch}. In general, normalization layers standardize its input $\bm{y}$,
as follows:

\begin{equation} \label{eq:norm} 
\bar{\bm{y}} = \frac{\bm{y} - \bm{\mu}}{\sqrt{\bm{\sigma^2} + \epsilon}} \times\bm{\gamma} + \bm{\beta}.
\end{equation}

where, the mean $\bm{\mu}$ and standard deviation $\bm{\sigma}$ are calculated across a specific dimension (batch, feature map, channel groups) depending on the type of normalization method. For instance, batch normalization (BN)~\cite{ioffe2015batch} normalizes activations across a mini-batch, layer normalization (LN)~\cite{ba2016layer} normalizes across all features of a single example, Instance Normalization (IN)~\cite{ulyanov2016instance} normalizes independently within each channel of a single example, and Group Normalization (GN)~\cite{wu2018group} normalizes across groups of channels. Furthermore, $\bm{\gamma}$  and $\bm{\beta}$ are learnable parameters and $\epsilon$ is a small constant for numerical stability.

\subsection{Model Ensemble and Related works}

As stated earlier, in the literature, several methods for uncertainty estimates are proposed. Among them, the model ensemble method is highly successful due to its high inference accuracy and quality uncertainty estimates. 

Model ensemble involves combining predictions from multiple individual models (see Fig.~\ref{fig:tiny_de_vs_related}(a)) to improve overall performance and estimate uncertainty. During training, $M$ models are trained independently or collaboratively using techniques such as bagging or boosting. These models can be trained with different architectures, initializations, or subsets of data to encourage diversity. During inference, predictions from different models are aggregated using methods such as averaging or weighted averaging to obtain the final prediction. Since training, storage, and processing of $M$ full models is required, the hardware cost, e.g., memory, latency, and power is a concern.

In Section~\ref{sec:intro}, related studies on model ensembles for uncertainty estimates and related works for cost reduction were discussed. Nevertheless, the ensemble of models has been extensively studied to improve model performance~\cite{hansen1990neural, dietterich2000ensemble, opitz1999popular}. Even in this case, there are several methods to reduce the cost of inference. For example, the work in~\cite{buciluǎ2006model} proposed a model compression technique to compress large and complex models into smaller and faster ones. Similarly,~\cite{hinton2015distilling} introduced the knowledge distillation method, which distills model ensembles into a single neural network. 

Since ensembles require training $M$ models, several studies aim to reduce their cost at training time. For example, \cite{huang2017snapshot}
proposed the Snapshot ensemble method, which encourages a single model to visit multiple local minima by training it using cyclic learning rates~\cite{loshchilov2016sgdr}. This method encourages the exploration of numerous local minima, which are then used as ensemble members.

In contrast, our approach aims to optimize performance, training, and inference costs collectively with AI accelerator architectures in mind.
\section{Tiny Deep Ensemble (Tiny-DE)} \label{sec:methodology}

As mentioned previously, a naive ensemble approach incurs significant memory and computational overhead. Here, the proposed Tiny Deep Ensemble approach is discussed, a low-cost ensemble method for uncertainty estimation in deep neural networks.

\subsection{Core Idea}

In Tiny-DE, only the normalization layers are ensembled, which overall have the smallest amount of parameters in the network, differ between the ensemble members, while all other weights are shared. We denote the normalization layers of layer index $l$ by $N_0^l, N_1^l, \dots, N_{M-1}^l$ in the following.
The normalization layers can be Batch Normalization, Layer Normalization, Instance Normalization, and Group Normalization with learnable parameters $\beta\in\mathbb{R}^{n}$ and $\gamma\in\mathbb{R}^{n}$. Therefore, compared to the deep ensemble approach~\cite{deep_ensemble} and BatchEnsemble~\cite{batchensemble}, our approach requires a $M\times$ lower weight matrix storage and a $2M\times$ lower rank-1 matrix computation (see Figs.\ref{fig:tiny_de_vs_related} (a) and (b)).  

\subsection{Operation Modes}
Depending on the batch processing capabilities of the hardware architecture, Tiny-DE can operate in either sequential or parallel modes. In hardware architectures where batch processing is challenging, such as the memristor-based computation-in-memory (CIM) architecture~\cite{hamdioui2015memristor, yu2021compute, mutlu2019processing}, a sequential processing NN architecture should be used. Here, "sequential" refers to sequential in time rather than signal flow through the ensembles.
In contrast, in parallel mode, \emph{single-shot uncertainty estimation} can be done using vectorization in hardware architectures such as edge tensor processing units (TPUs), field-programmable gate arrays (FPGAs), and graphics processing units (GPUs)~\cite{jouppi2017datacenter, posewsky2016efficient}. Both methods are described in detail in the following.

\subsubsection{\textbf{Sequential Inference}}

The sequential inference of Tiny-DE utilizes a counter variable $c$ and router to dynamically select a normalization layer for each forward pass. Depending on the state of the counter $c$, the output of the $l$-th layer $y^l$ is directed through one of the $M$ normalization layers, as shown in Fig.~\ref{fig:tiny_de_vs_related} c). The activation function such as the ReLU function is applied to the processed output as is normally done. 



The counter $c$ is an unsigned integer and 
it is updated cyclically in each layer as follows:
\begin{equation}
c \gets (c + 1) \mod M,
\end{equation}
where $c$ is initialized to $0$ at the start of the inference process. The mechanism ensures that the output of each layer sequentially passes through each normalization layer in a cyclic order. For example, if $c=0$, the output $y^l$ is processed by $N_0$. In the next forward pass, $c$ becomes $1$, routing the output $y^l$ through $N_1$, and this process is repeated until the $M$-th forward pass. After that, $c$ resets to $0$. Note that, due to the global signal routing and synchronization challenge, the counter variable is updated locally in each layer. 

This cyclic routing mechanism allows each input of the NN to experience every normalization setting, providing diverse internal-state manipulations within a single inference cycle, which is crucial for enhancing the ensemble's ability to generalize and generate output distribution for uncertainty estimation.

\begin{figure}
    \centering
    \includegraphics[width=\linewidth]{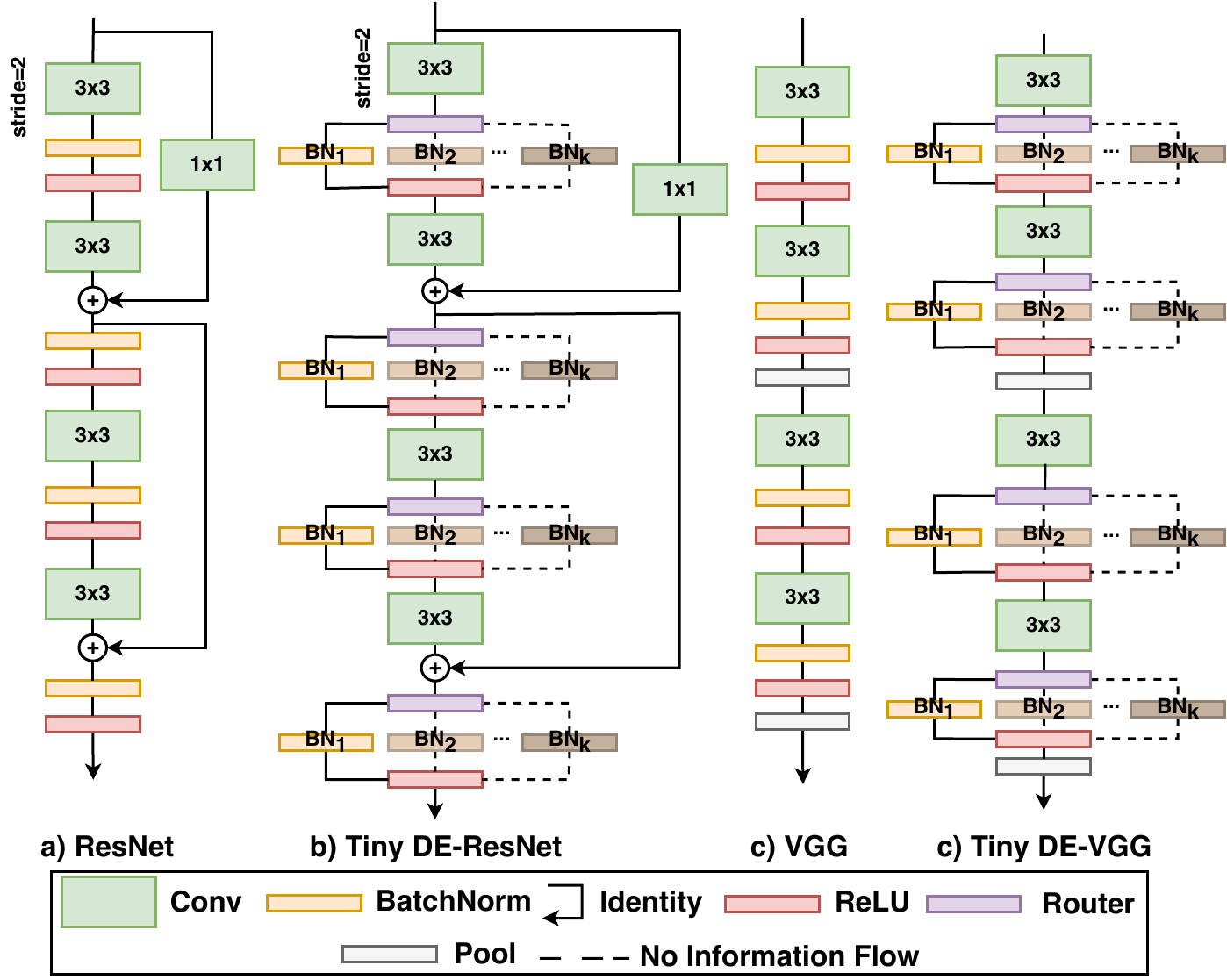}
    \caption{Sketch of proposed Tiny-DE architecture based on popular CNN architectures ResNet~\cite{he2016deep} and VGG~\cite{simonyan2014very}. We only show the four signature layers of a specific topology. Our proposed topology is generalizable across existing topologies, with only the addition of a router before the normalization layers. In the case of our proposed approach in batch mode, no change is required in the topology.}
    \label{fig:topology}
\end{figure}

Furthermore, the proposed Tiny-DE can be generalized to all existing NN architectures by making minor modifications, as shown in Fig.~\ref{fig:topology}. For popular architectures, such as ResNet and VGG, a router can be inserted after the convolutional layer. 

\paragraph*{\textbf{Router Implementation}}
In CIM architectures, the router can be implemented digitally at the periphery using a demultiplexer (DeMux). The DeMux takes the $Q$-bit unsigned counter $c$ as the control signal, allowing for up to $2^Q$ possible routing paths, each corresponding to one of the normalization layers (ensemble members). Since a typical DeMux expects a bit-wise control signal, the DeMux for our purpose is designed to interpret the control signal $c$ in binary representation. This can be expressed as:
\begin{equation}
\operatorname{binary}(c) = b_{Q-1} b_{Q-2} \cdots b_0,
\end{equation}
where $b_{Q-1}$ to $b_0$ are the bits of the binary sequence representing $c$. 

Our approach requires only changes to the CiM periphery since the router is implemented in the digital domain with some logic hardware. Specifically, the Multiply-Accumulate (MAC) operation of a layer is computed in a memristor-based crossbar structure (analog domain) and the result is digitized by an analog-to-digital converter (ADC) operation. Following that the router selects the parameter for normalization and the normalization is performed. In the following, non-linear activation is performed and a digital-to-analog (DAC) converts the results of the activation function for MAC operation (in the analog domain) of the subsequent layer. 
The overall algorithm for our proposed approach in sequential inference mode is depicted in Algorithm~\ref{algo:router}.

\begin{algorithm}
\caption{Sequential inference mode of Tiny-DE in CiM}
\begin{algorithmic}[1]
\State \textbf{Input:} Controller $c$, number of ensembles $M$, input to the network $\bm{x}$, number of layers $L$


\For{$m=1, \dots, M$} \Comment{sequential inference}
    \For{$l=1, \dots, L$} \Comment{single forward pass}
    \State Digital-to-analog conversion
    \State MAC operation in memristor-based crossbar array
    \State Analog-to-digital conversion
    \State Router selects parameters of normalization layer
    \State Perform normalization
    \State Non-linear activation
    \EndFor
\State Increment counter
\EndFor
\end{algorithmic}
\label{algo:router}

\end{algorithm}
\vspace{-.4em}

\subsubsection{\textbf{Single-Shot Uncertainty Estimation}}
By manipulating the computations for a mini-batch, the computations of the Tiny-DE approach are parallelizable within a hardware architecture that allows batched processing such as FPGAs, GPUs, and TPUs. Therefore, only a \emph{single forward pass} with respect to multiple ensemble members in parallel is required to estimate uncertainty. Here, an input to the convolution or linear layer is repeated $M$ times to generate a mini-batch of size $M$ to obtain the batched output $\bm{Y}^l$. However, if the batch size of the inference inputs is more than one, e.g., $B$, by repeating the input similarly $M$ times, an effective batch size of $M \cdot B$ can be created. Therefore, a single forward pass is required for the convolution or linear layer.

However, to still allow a single forward pass through all ensemble members, we propose \emph{EnsembleNorm}. 
In EnsembleNorm, the input dimension and the parameters are modified across the batch dimension so that they independently apply normalization to each input of the batch. Specifically, the input of the shape $[M \cdot B, C, H, W]$ is reshaped as $[M, B, C, H, W]$. Here, $C$, $H$, and $W$ represent the channel, height, and width, respectively. Similarly, the learnable parameters expanded to $\beta\in\mathbb{R}^{M\times n}$ and $\gamma\in\mathbb{R}^{M\times n}$. That means that the parameters are not only channel-specific, but also unique to each ensemble member. The mean and variance are also calculated in the respective dimensions. That means that each ensemble member can have its own specific mean $\bm{\mu}_{m,c}$ and variance $\bm{\sigma}^2_{m,c}$. Furthermore, each ensemble member can be scaled and shifted by its own unique parameters, $\bm{\gamma}_{m,c}$ and $\bm{\beta}_{m,c}$.





Subsequently, the normalized output $\bar{\bm{Y}}^l$ is reshaped again to $[M \cdot B, C, H, W]$ before applying the non-linear activation function. The PyTorch implementation of EnsembleNorm with other implementations will appear in~\footnote{will be open-sourced upon acceptance}.


Consequently, all ensemble members can compute the output in a single forward pass, eliminating the need to calculate the output of each ensemble member sequentially. Therefore, the computational latency is reduced to a minimum.

\subsection{Training}
The training procedure of Tiny-DE also depends on the operating mode. The sequential mode involves two main phases, but the parallel mode allows single-shot training. Both methods are described in detail in the following.

\paragraph*{\textbf{Sequential Mode}}

As stated earlier, the overall training of the $M$ ensembles requires two main phases. Initially, the full model is trained with all parameters (weights and biases) being updated. After this, the parameters of the model, e.g., weights and biases are frozen, and the normalization layers are re-initialized. Here, "frozen" means that they are not updated using backpropagation. In each subsequent training, only the normalization layers are updated. The training is stopped once a comparable accuracy to the full model is achieved. All trained parameters of the normalization layer are accumulated in a list to allow for ensemble learning as described earlier.

Since the full model is only trained once, the training overhead and complexity are significantly lower compared to~\cite{deep_ensemble} and~\cite{batchensemble}, respectively. The decoupling of parameters allows for effective ensemble learning without the overhead of training multiple distinct models from scratch. In addition, it allows one to \emph{obtain $M$ ensemble members from a single pre-trained model}.

\paragraph*{\textbf{Single-Shot Training}}
In the batched processing mode, replacing the normalization layer with the proposed EnsembleNorm layers along with manipulating the dimension as discussed earlier section, all the ensemble members can be trained together. 

Here, the effective batch size for training may need to be reduced due to the memory overflow issue in GPUs. However, since training is typically done in the cloud, it is not an issue for edge inference.

\subsection{Prediction and Uncertainty Estimation}

The input for inference is forward-passed through the Tiny-DE to get the predictive distribution. The final prediction of Tiny-DE is obtained from the average predictions of all ensemble members. 

To obtain uncertainty in the prediction, we explore different methods depending on the task. For classification tasks, the predictive entropy is commonly used, but we also measure the maximum disagreement among the outputs, as shown in Algorithm~\ref{algo:max_diss}.
\begin{algorithm}
\caption{Maximum Disagreement}
\begin{algorithmic}[1]
\State \textbf{Input:} output samples of $\bm{y}$ of shape $(M, B, K)$
\State Initialize Max Disagreement (MD) with zeros of shape $(B, K)$
\For{$m = 1, \dots, M-1$}
    \For{$m' = m+1, \dots M$}
        \State Calculate absolute difference $m'$ and $m$ output 
        \State Calculate the maximum across the class dimension
        \State Update Max Disagreement
    \EndFor
\EndFor
\end{algorithmic}
\label{algo:max_diss}
\end{algorithm}

The Maximum Disagreement metric quantifies uncertainty by calculating the maximum absolute difference in output distributions for each class, across all models in the ensemble. Since it is computed directly from SoftMax output, this metric ranges from 0 to 1. A low maximum disagreement value (closer to 0) indicates low uncertainty, and a high value (closer to 1) indicates high uncertainty.

Furthermore, in semantic segmentation and time series prediction tasks, uncertainty is quantified by the variance in 
predictions of different ensemble members. Lastly, for regression tasks, the uncertainty is estimated using the negative log-likelihood (NLL) of the prediction.


\subsection{Diversity Improvement Among Ensemble Members}\label{sec:diversity}
Diverse predictions among ensemble members are advantageous as they offer complementary perspectives, potentially improving performance and enhancing uncertainty estimates. For our approach, diversity can be improved by a) using different kinds of normalization layers in each member, b) training each ensemble member with different data augmentations, and c) creating multiple bootstrap samples (random samples with replacement) from the training data and training each ensemble member on each sample.

\section{Results}\label{sec:experiments}

\subsection{Experimental Setup}

To show scalability on deep learning tasks, we have evaluated our method on four different tasks: image classification, regression, autoregressive time series forecast, and semantic segmentation. To further show scalability on datasets and NN topologies, we have evaluated each task on several state-of-the-art (SOTA) NN topologies (including CNN and RNN) and datasets.

For image classification, we used the CIFAR-10 and CIFAR-100 benchmark datasets on the VGG-19, ResNet-56, ShuffleNet-V2, RepVGG-A1, and TinyML compatible MobileNet-V2 CNN topologies. Furthermore, for the regression task, we have used 10 UCI datasets with a topology and setting as~\cite{gal2016dropout}. Specifically, each dataset except for the protein and Year Prediction MSD, is split into 20 train-test folds. Five train-test splits were used for the protein dataset, and a single train-test split was used for the Year Prediction MSD dataset. The NN has 2-hidden layers with ReLU6 nonlinearity followed by a 1D batch normalization layer. The number of neurons is 50 for the smaller datasets and 100 for the larger protein and Year Prediction MSD datasets, making the network compatible with edge AI accelerators. All the dataset was trained for 40 epochs and we have used 5 ensemble members (M=5).

On the other hand, for the time-series forecast, an NN with an LSTM layer and a classifier layer was used for the Mauna Loa CO2 concentrations dataset. Lastly, for Semantic segmentation tasks, we have considered binary as well as multi-class segmentation datasets and two safety-critical scenarios, for biomedical and automotive. For biomedical image segmentation, we have used the Kvasir-SEG~\cite{jha2020kvasir} dataset which contains medically obtained gastrointestinal polyps images on the Feature Pyramid Network (FPN)~\cite{lin2017feature}. For automotive scene understanding we used the CamVid~\cite{brostow2009semantic} dataset which consists of road scene images and involves segmenting each pixel into one of the 12 classes on the UNet++ topology~\cite{zhou2019unetpp}. We have further evaluated the generalized scene understanding task with the Pascal VOC dataset with the fully convolutional network (FCN). The encoder network for each topology is shown in brackets in Table.~\ref{tab:segments}. 

Note that the semantic segmentation task is known to be more challenging than other tasks due to its finer granularity. That is, it involves segmenting an image into multiple sections and assigning each pixel with its corresponding class label.

The performance of the classification task is evaluated on inference accuracy, time series, and regression on root-mean-square-error (RMSE), and semantic segmentation on pixel accuracy and mean intersection-over-union (mIoU) metrics. 

In terms of uncertainty estimation, classification tasks are evaluated on data distribution shift and out-of-domain data as OoD data. Specifically, for data distribution shift, images are corrupted by $90^\circ$ rotation and Gaussian noise, a subset of the CIFAR-C dataset~\cite{hendrycks2019benchmarking}. Furthermore, SVHN (Street View House Numbers) and STL-10 datasets are used for out-of-domain data which refers to data that significantly deviates from the distribution of the training data. The predictive entropy distribution is calculated from the mean of 250 batch samples and is subsequently modeled as a normal distribution.

\newcommand{\tpm}{$\pm$}
\begin{table*}[t!]
\caption{Results on regression benchmark datasets of the proposed approach and related works Probabilistic back-propagation (PBP)~\cite{hernandez2015probabilistic}, MC-Dropout~\cite{gal2016dropout}, Deep Ensembles~\cite{deep_ensemble} comparing RMSE and NLL. Dataset size ($N$) and input dimensionality ($Q$) are also given.}
\vspace{-1em}
\center
\footnotesize
\resizebox{\linewidth}{!}{
\begin{tabular}{@{}l@{\hspace{4mm}}l@{\hspace{3mm}}l@{\hspace{4mm}}r@{\hspace{2mm}}r@{\hspace{2mm}}r@{\hspace{4mm}}r@{\hspace{2mm}}r @{\hspace{2mm}}r@{\hspace{2mm}}r @{\hspace{2mm}}r @{\hspace{2mm}} r}  
\multicolumn{3}{c}{} & 
\multicolumn{4}{c}{\footnotesize Avg. Test RMSE and Std. Errors $\downarrow$} & 
\multicolumn{4}{c}{\footnotesize Avg. Test LL and Std. Errors $\downarrow$} \\ 
\textbf{Dataset} & $N$ & $Q$ & 
\multicolumn{1}{c}{\textbf{PBP}} & 
\multicolumn{1}{c}{\textbf{MC-Dropout}} & 
\multicolumn{1}{c}{\textbf{Deep Ensemble}} & 
\multicolumn{1}{c}{\textbf{Proposed}} & 
\multicolumn{1}{c}{\textbf{PBP}} & 
\multicolumn{1}{c}{\textbf{MC-Dropout}} & 
\multicolumn{1}{c}{\textbf{Deep Ensemble}} & 
\multicolumn{1}{c}{\textbf{Proposed}} & \\ 
\hline 
Boston Housing & 506 & 13 & {3.01 $\pm$ 0.18}  & {2.97 $\pm$ 0.85}  & {3.28 $\pm$ 1.00} & \textbf{2.97 \tpm 0.46} & {2.57 $\pm$ 0.09}  & \textbf{2.46 $\pm$ 0.25}  & \textbf{2.41 $\pm$ 0.25} & {4.92 \tpm 1.03} \\
Concrete Strength & 1,030 & 8 & {5.67 $\pm$ 0.09}  & {5.23 $\pm$ 0.53}  & {6.03 $\pm$ 0.58} & \textbf{5.51 \tpm 0.41} & {3.16 $\pm$ 0.02}  & \textbf{3.04 $\pm$ 0.09}  & {3.06 $\pm$ 0.18} & {5.02 \tpm 0.62} \\ 
Energy Efficiency & 768 & 8 & {1.80 $\pm$ 0.05}  & {1.66 $\pm$ 0.19}  & {2.09 $\pm$ 0.29} & \textbf{1.53 \tpm 0.38} & 2.04 $\pm$ 0.02  & 1.99 $\pm$ 0.09  & \textbf{1.38 $\pm$ 0.22} & \textbf{1.41 \tpm 0.46} \\ 
Kin8nm & 8,192 & 8 & 0.10 $\pm$ 0.00  & 0.10 $\pm$ 0.00  & {0.09 $\pm$ 0.00} & \textbf{0.07 \tpm 0.00} & -0.90 $\pm$ 0.01  & -0.95 $\pm$ 0.03  & \textbf{-1.20 $\pm$ 0.02}  & {-0.95 \tpm 0.01} \\ 
Naval Propulsion & 11,934 & 16 & 0.01 $\pm$ 0.00  & 0.01 $\pm$ 0.00  & {0.00 $\pm$ 0.00} & \textbf{0.00 \tpm 0.00} & -3.73 $\pm$ 0.01  & -3.80 $\pm$ 0.05  & \textbf{-5.63 $\pm$ 0.05} & {-3.81 \tpm 0.08} \\ 
Power Plant & 9,568 & 4 & {4.12 $\pm$ 0.03}  & \textbf{4.02 $\pm$ 0.18} & {4.11 $\pm$ 0.17} &  {4.48 \tpm 0.18} & 2.84 $\pm$ 0.01  & {2.80 $\pm$ 0.05}  & \textbf{2.79 $\pm$ 0.04} & {2.95 \tpm 0.05} \\ 
Protein Structure & 45,730 & 9  & 4.73 $\pm$ 0.01  & {4.36 $\pm$ 0.04}  & 4.71 $\pm$ 0.06  & \textbf{3.92 \tpm 0.03} & 2.97 $\pm$ 0.00  & 2.89 $\pm$ 0.01  & \textbf{2.83 $\pm$ 0.02} & 5.05 \tpm 0.52 \\ 
Wine Quality Red & 1,599 & 11 & {0.64 $\pm$ 0.01}  & {0.62 $\pm$ 0.04}  & {0.64 $\pm$ 0.04} & \textbf{0.64 \tpm  0.05} & 0.97 $\pm$ 0.01  & {0.93 $\pm$ 0.06}  & {0.94 $\pm$ 0.12} &{1.28 \tpm 0.33} \\
Yacht Hydrodynamics & 308 & 6 & \textbf{1.02 $\pm$ 0.05}  & {1.11 $\pm$ 0.38}  & {1.58 $\pm$ 0.48} & 3.22 \tpm 1.59 & 1.63 $\pm$ 0.02  & 1.55 $\pm$ 0.12  & \textbf{1.18 $\pm$ 0.21} & 1.37 \tpm 0.43 \\ 
Year Prediction MSD & 515,345 & 90 & 8.88 $\pm$ NA  & {8.85 $\pm$ NA}  & 8.89 $\pm$ NA & \textbf{8.53 \tpm NA}  & 3.60 $\pm$ NA  & 3.59 $\pm$ NA  & \textbf{3.35 $\pm$ NA} & 7.63 $\pm$ NA\\ 
\hline 
\end{tabular} 
}
\label{table:regression}
\vspace{-2em}
\end{table*}

\subsection{Evaluation of Regression on Real-World UCI Datasets}

The result of the regression task is depicted in Table~\ref{table:regression}. Our approach is compared with Bayesian~\cite{hernandez2015probabilistic}, implicit ensemble (MC-Dropout)~\cite{gal2016dropout}, and ensemble~\cite{deep_ensemble} methods. As can be seen, our method outperforms or is competitive with existing methods in terms of RMSE and NLL. Specifically, our method outperforms other methods in 8 out of the 10 datasets in terms of RMSE. In some datasets, we observe that our method is slightly worse in terms of NLL. 
We believe that this is due to the fact that our method optimizes for RMSE instead of NLL (which captures predictive uncertainty). We found that there is a trade-off between RMSE and NLL. Optimizing for NLL instead reduces RMSE. Also, we did not perform hyperparameters optimization, unlike~\cite{gal2016dropout} which performed grid search. 

\subsection{Evaluation of Classification}

In classification tasks with various topologies, it can be observed that \ourmethod improves inference accuracy by up to $0.81\%$ or is comparable with the single model, as shown in Table~\ref{tab:classification_results}.

\begin{figure}
    \centering
    \includegraphics[width=0.86\linewidth]{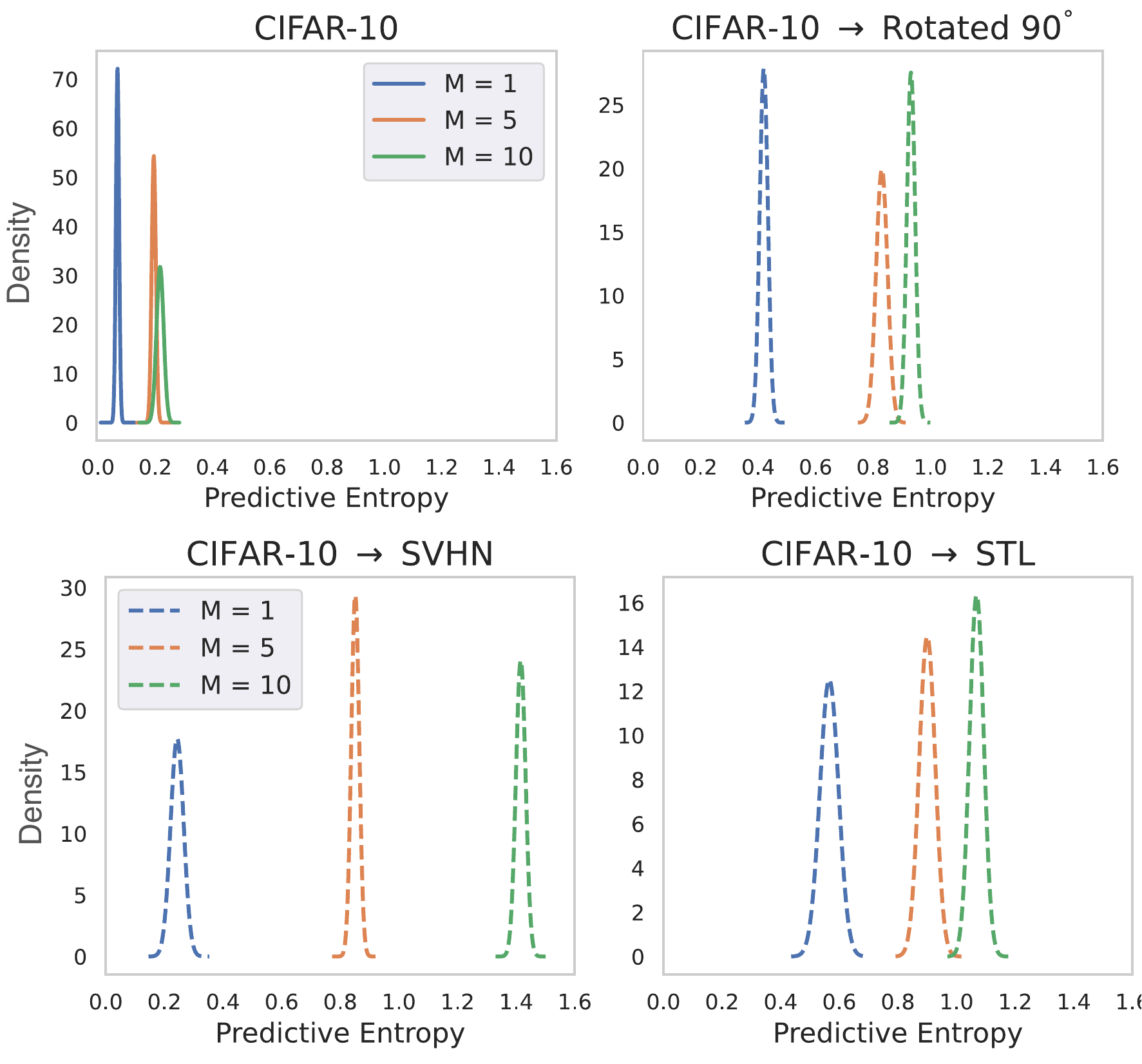}
    \caption{Uncertainty distributions for the Tiny-DE approach on CIFAR-10, including ID CIFAR-10, and OOD datasets such as rotated CIFAR-10, SVHN, and STL. Notably, larger ensembles show increased relative change of uncertainty distribution from ID compared to a single model (M = 1).
    }
    \label{fig:CIFAR_10_Uncer}
    \vspace{-1.2em}
\end{figure}

In terms of uncertainty estimates in the OoD data, Fig.~\ref{fig:CIFAR_10_Uncer} shows the predictive uncertainty of the ResNet-32 model trained on clean
CIFAR-10.
It can be observed that the predictive entropy is low in clean CIFAR-10, that is, ID data. However, if the model receives OoD data, e.g., rotated, SVHN, or STL-10 data, the predictive entropy increases from baseline. Importantly, the relative change in the predictive entropy is significantly higher for our proposed Tiny-DE approach. Here, the relative change in the uncertainty estimates signifies better capabilities in the uncertainty estimates. Furthermore, the change in predictive entropy becomes greater as the number of ensembles increases, which is an ideal behavior.

\begin{figure}
    \centering
    \includegraphics[width=\linewidth]{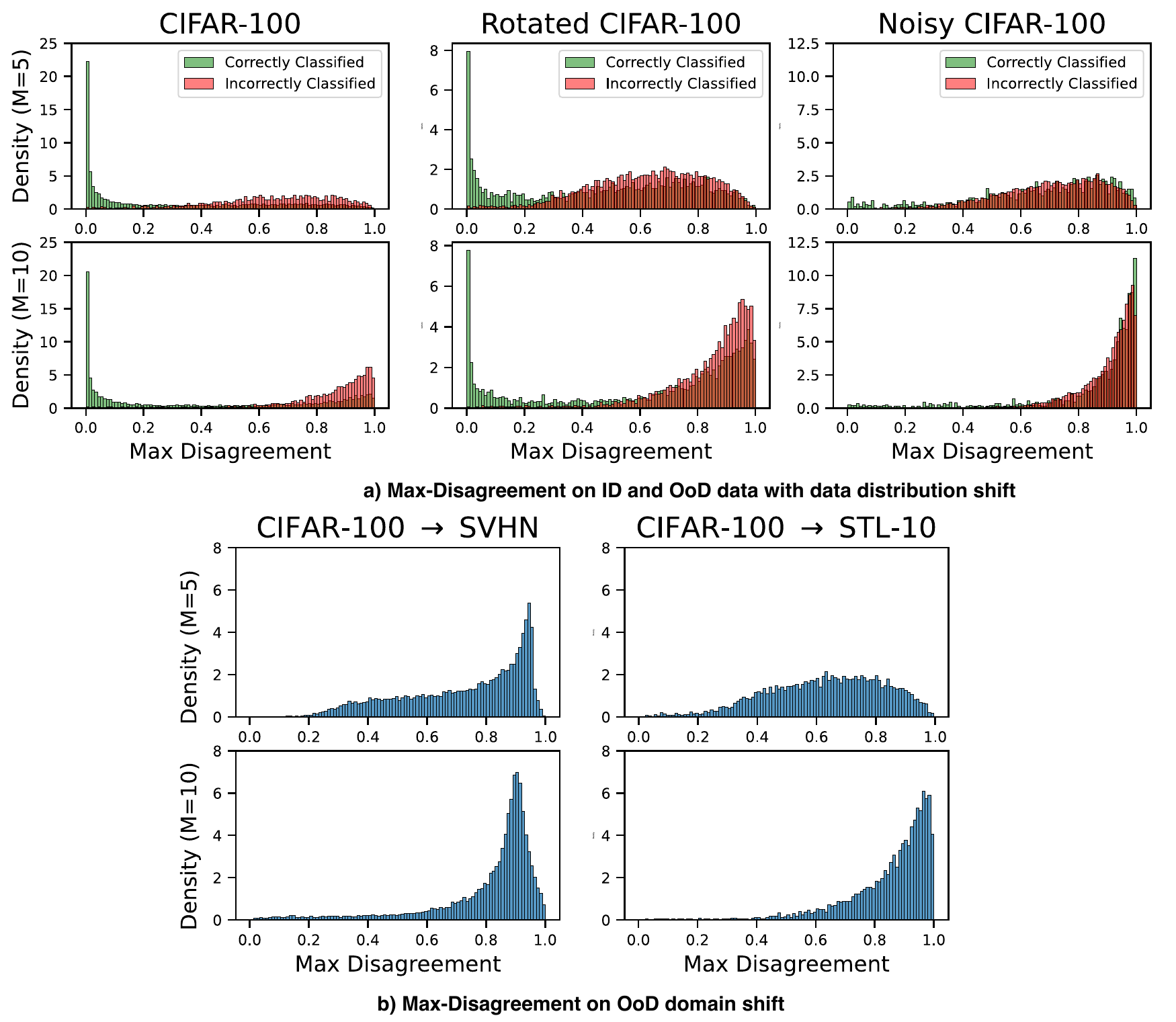}
    \caption{ID and OoD Max Disagreement distributions for the Tiny-DE approach trained on clean CIFAR-100 (ID).
    Notably, larger ensembles show increased relative change of uncertainty distribution from ID. 
    }
    \label{fig:CIFAR_100_Uncer}
\end{figure}

In contrast, the CIFAR-100 model is evaluated on the max disagreement metric, as shown in Fig.~\ref{fig:CIFAR_100_Uncer}. In ID data, our approach shows finer granularity in uncertainty estimates. Specifically, the uncertainty is low for correctly classified images and high for incorrectly classified images. On corrupted (rotated and noisy) images, OoD data, the model can still predict some images correctly. Our approach shows a similar uncertainty distribution for correctly and incorrectly predicted images. In addition, the relative change from the baseline distribution is also high. In domain-changed data (SVHN and STL-10), our approach shows high uncertainty with distributions concentrated toward the right. Furthermore, the distributions shift more toward the right as the number of ensembles increases.

\begin{table}[]
\centering
\footnotesize
\caption{Performance of Tiny-DE with CIFAR-10 and CIFAR-100 dataset trained on various topologies with up to 15 ensemble members.}

\begin{tabular}{cccccc}
\multirow{2}{*}{Topology} & \multirow{2}{*}{Dataset}                    & \multicolumn{4}{c}{Number of ensembles}                                                        \\ \cline{3-6} 
                          &                                                & 1                    & 5                    & 10                   & 15                   \\ \hline
VGG-19                    & \multirow{5}{*}{CIFAR-10}                      & 93.91                    & 93.86                    & 93.79                    & 93.80                    \\
ResNet-56                 &                                               &         94.37             &        94.28              &               94.14       &       94.38               \\
ShuffleNet-V2             &                                                & 93.3                    & 93.27                    & 93.44                    &    93.67                 \\
RepVGG-A1                 &                                                & 94.93                    & 94.56                    & 94.84                    & 94.62                    \\
MobileNet-V2              &                                               & 94.05                   & 93.67                     & 93.92                    & 94.01                    \\ \hline
VGG-19                    & \multicolumn{1}{l}{\multirow{5}{*}{CIFAR-100}} & \multicolumn{1}{l}{73.87} & \multicolumn{1}{l}{74.21} & \multicolumn{1}{l}{74.56} & \multicolumn{1}{l}{74.68} \\
ResNet-56                 & \multicolumn{1}{l}{}                           & \multicolumn{1}{l}{72.63} & \multicolumn{1}{l}{72.64} & \multicolumn{1}{l}{72.85} & \multicolumn{1}{l}{72.82} \\
ShuffleNet-V2             & \multicolumn{1}{l}{}                           & \multicolumn{1}{l}{72.58} & \multicolumn{1}{l}{72.75} & \multicolumn{1}{l}{73.54} & \multicolumn{1}{l}{73.11} \\
RepVGG-A1                 & \multicolumn{1}{l}{}                           & \multicolumn{1}{l}{76.44} & \multicolumn{1}{l}{75.77} & \multicolumn{1}{l}{74.67} & \multicolumn{1}{l}{75.21} \\
MobileNet-V2              & \multicolumn{1}{l}{}                           & \multicolumn{1}{l}{74.29} & \multicolumn{1}{l}{74.41} & \multicolumn{1}{l}{74.67} & \multicolumn{1}{l}{75.21} \\ \hline
\end{tabular}
\label{tab:classification_results}
\end{table}

\subsection{Evaluation of Time-Series Prediction}

\begin{figure}
    \centering
    \includegraphics[width=\linewidth]{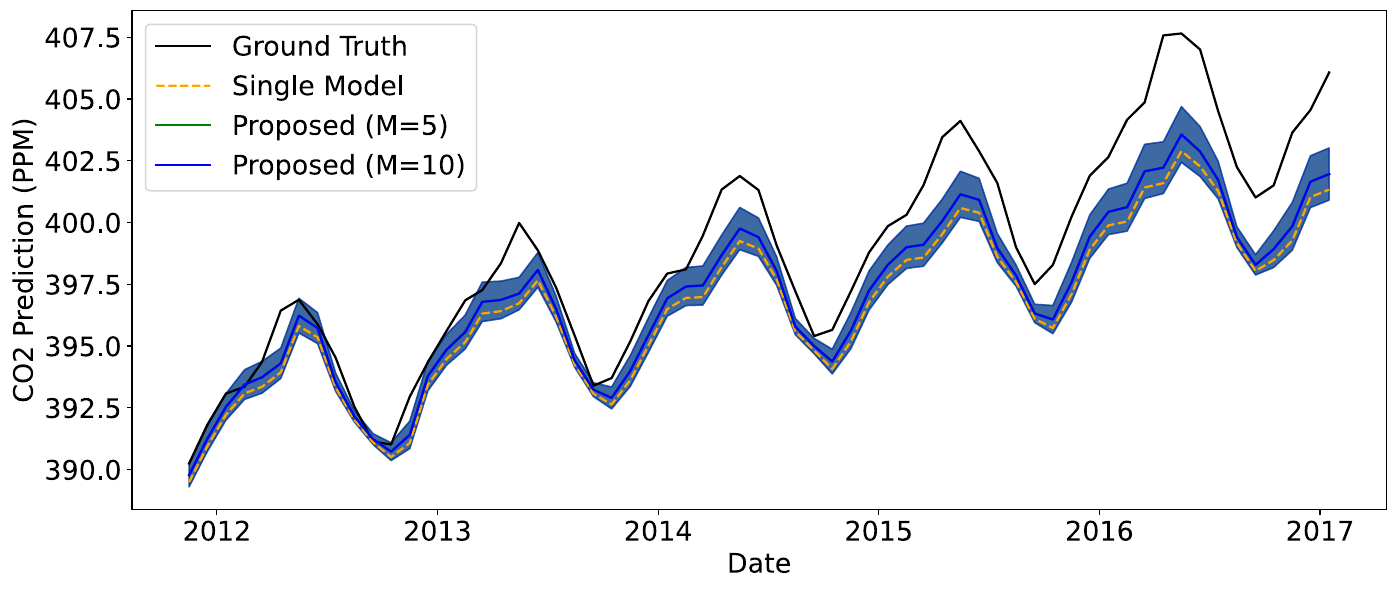}
    \caption{Auto-regressive time series prediction of atmospheric CO2 of a single model and our proposed Tiny-DE model with up to 10 ensemble members. The shaded region shows the uncertainty around prediction. }
    \label{fig:timeseries_co2}
    \vspace{-1em}
\end{figure}

The performance of our proposed approach on autoregressive time series prediction is shown in Fig.~\ref{fig:timeseries_co2}. As can be seen, the prediction curve is closer to the ground truth for our approach compared to the single model. Furthermore, the curve approaches ground truth as the number of ensemble members increases. Specifically, the single model achieves an RMSE score of $0.1119$. In contrast, our proposed Tiny-DE method achieves an RMSE score of $0.0943$ for 5 ensemble members, which is reduced to $0.0921$ for 10 members. That translates into a $17.7\%$ reduction in the RMSE score. In general, all models follow the same trend as the ground truth.

\subsection{Evaluation of Semantic Segmentation}

\begin{figure}
    \centering
    \includegraphics[width=\linewidth]{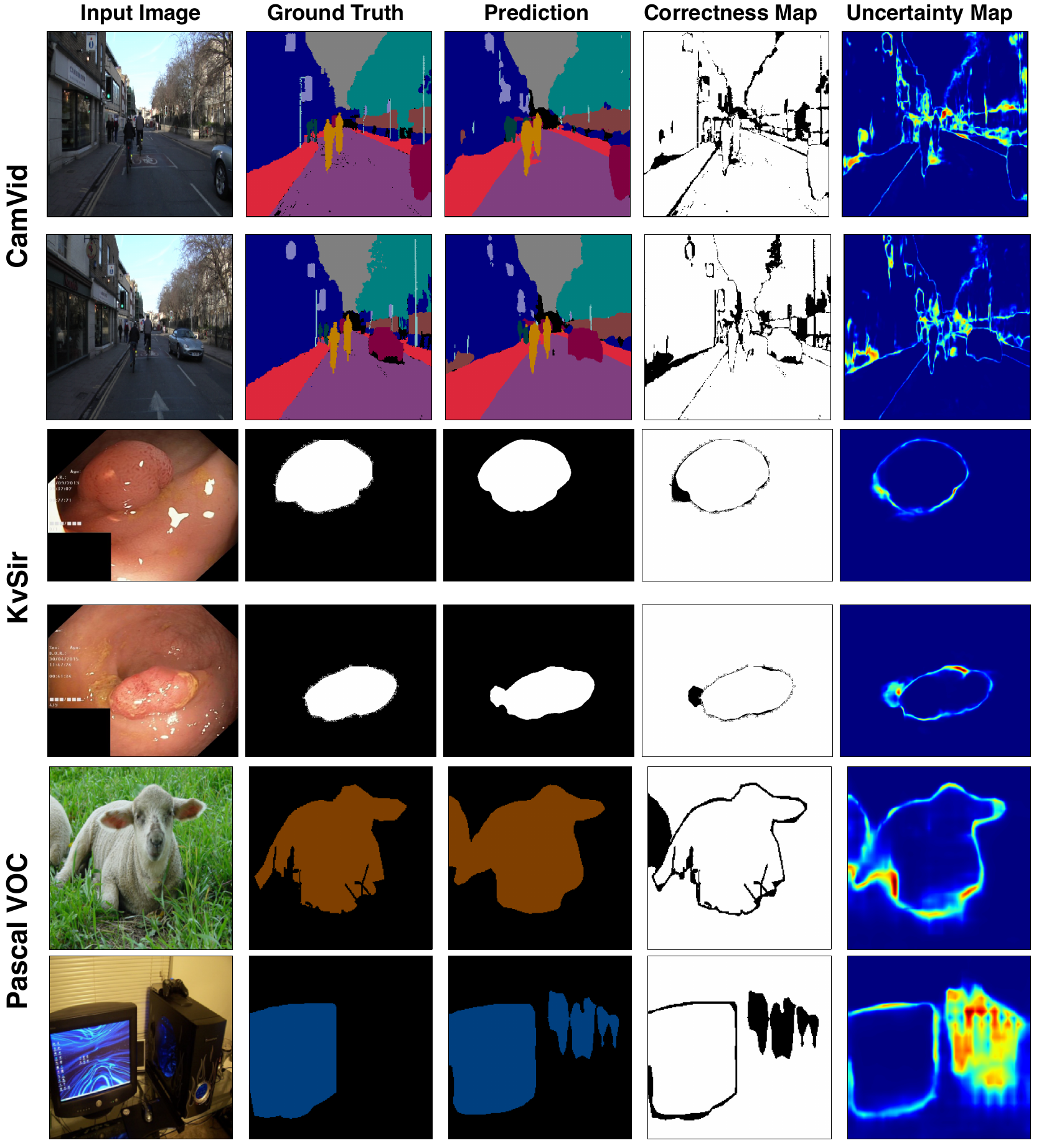}
    \caption{Qualitative results for several semantic segmentation tasks and associated uncertainty estimates. The correctness map is a binary diagram indicating correct and incorrect predictions in white and black, respectively. 
}
    \label{fig:segments}
    \vspace{-1em}
\end{figure}

Similarly, in semantic segmentation tasks with several challenging datasets and SOTA models, our approach performs comparably or outperforms the baseline model, as shown in Table~\ref{tab:segments}. Two qualitative examples of each dataset are shown in Fig.~\ref{fig:segments}. As can be seen, the predictions are close to the ground truth, with only incorrect predictions around the edges of segments or in uncommon classes. Here, uncommon classes refer to classes that occur infrequently or are less represented in the dataset. 

In terms of uncertainty estimates, our proposed approach can estimate uncertainty accurately. In an ideal case, misclassified pixels should have high uncertainty around them, and correctly classified pixels should have low uncertainty. As shown in Fig.~\ref{fig:segments} our approach captured this behavior effectively. 

\begin{table}[]
\caption{Pixel accuracy and mean intersection over union (IoU) of the single model and our proposed Tiny-DE (M = 5) with different datasets and SOTA models. }
\resizebox{\linewidth}{!}{
\begin{tabular}{cccccc}
Topology           & Dataset  & \multicolumn{2}{c}{Single Model} & \multicolumn{2}{c}{Proposed (M=5)} \\ \hline
                   &          & Pixel Acc         & mIoU         & Pixel Acc          & mIoU          \\ \cline{3-6} 
UNet++ (ResNet-34) & CamVid   & 91.65             & 63.95        & 91.52              & 63.99         \\
FPN (ResNet-18)    & KvaSir   & 95.95             & 74.62        & 95.89              & 74.57         \\
FCN (ResNet-50)    & CIFAR-10 & 87.78             & 69.63        & 87.71              & 68.58        
\end{tabular}
}
\label{tab:segments}
\end{table}

\subsection{Comparison with Related Works}

\begin{figure}
    \centering
    \includegraphics[width=\linewidth]{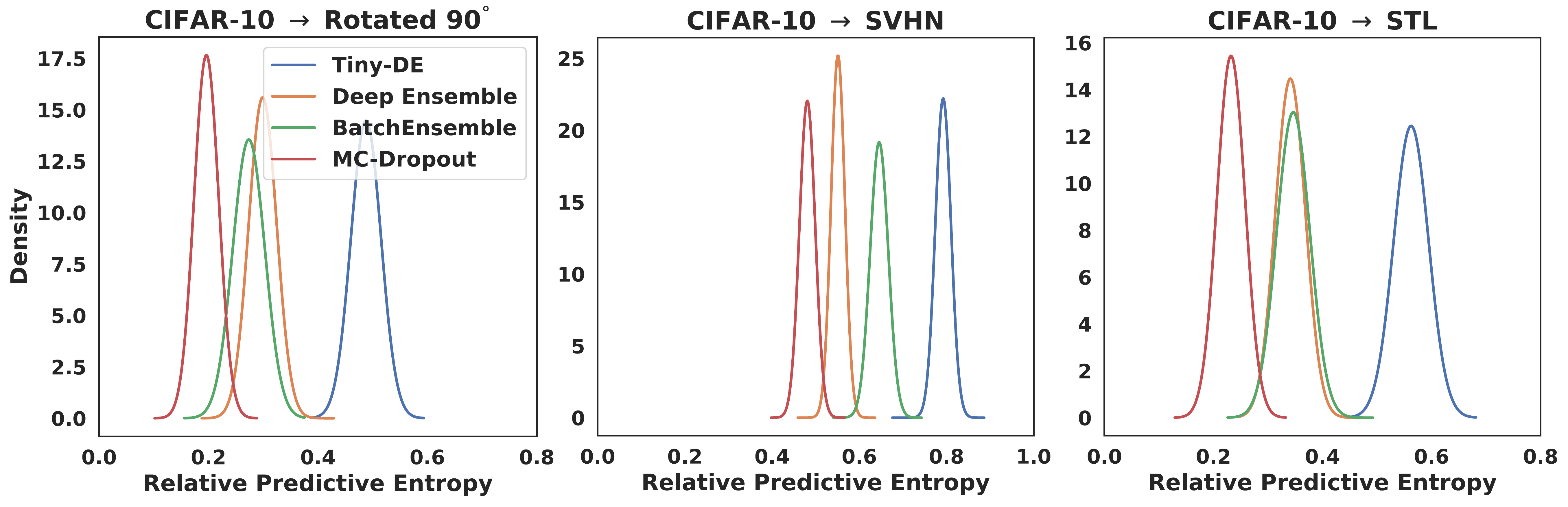}
    \caption{Relative change in predictive entropy on OoD data of Tiny-DE (ours) in comparison to Deep Ensemble~\cite{deep_ensemble}, MC-Dropout~\cite{gal2016dropout}, and BatchEnsemble~\cite{batchensemble}.}
    \label{fig:enropy_related}
\end{figure}

In the presence of OoD data, the higher the relative change in predictive entropy with respect to ID distribution, the better the method. Compared to related uncertainty estimation methods with model ensemble~\cite{deep_ensemble, gal2016dropout, batchensemble}, the relative predictive entropy of our Tiny-DE approach is much higher, as shown in Fig.~\ref{fig:enropy_related}. This further underscores the robustness of our approach. Here, the validation is done on the ResNet-32 topology on the CIFAR-10 dataset, but we found that this translates to other topologies and datasets. 

\subsection{Improving Diversity}

\begin{figure}
    \centering
    \includegraphics[width=\linewidth]{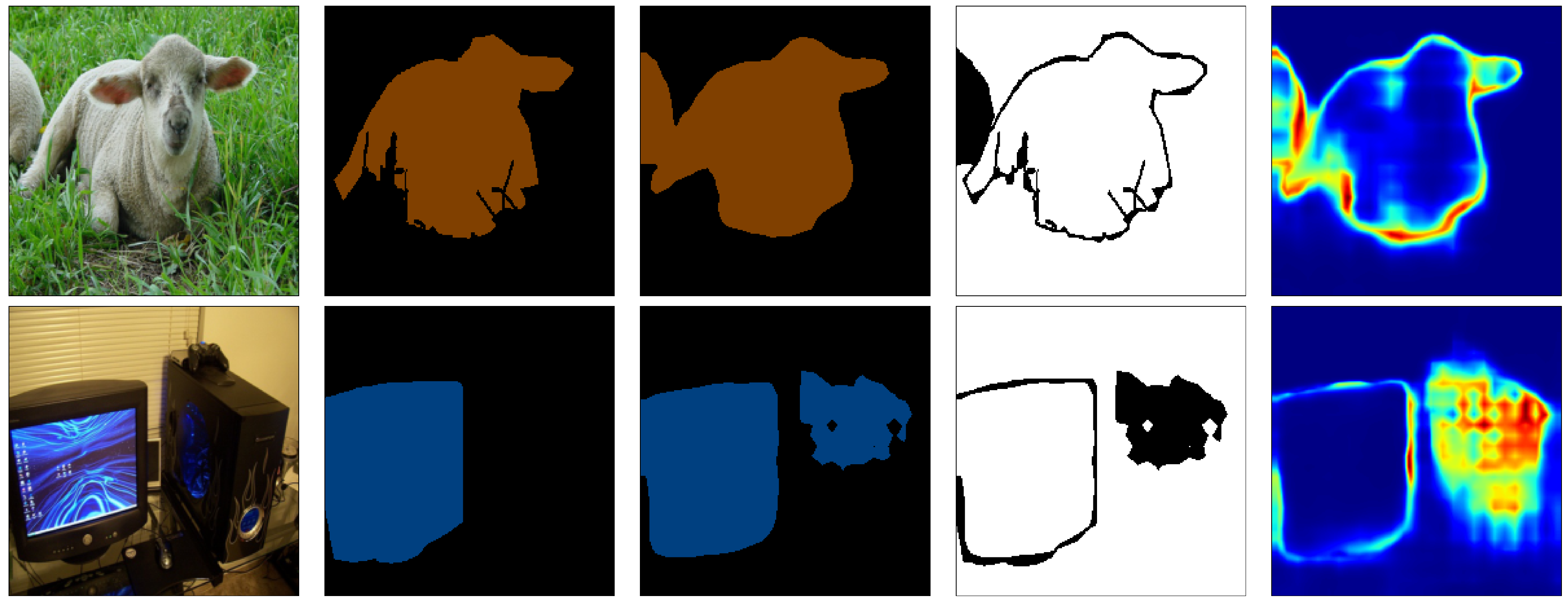}
    \caption{Results for Pascal VOC with improved diversity in ensemble members using different random data augmentation.}
    \label{fig:diversity}
    \vspace{-1em}
\end{figure}

As mentioned in Section~\ref{sec:diversity}, more diversity among the prediction of the ensembling members can lead to better performance and uncertainty estimates.
Therefore, we have performed another set of experiments in which each ensemble member is trained with different data augmentations. We found that by improving diversity with different random data augmentation to train each ensemble member, the uncertainty estimates increase on OoD data. For example, as shown in Fig.~\ref{fig:diversity} the uncertainty maps around incorrect pixels become stronger compared to Fig.~\ref{fig:segments} when each ensemble member is trained using different data augmentations. Furthermore, pixel accuracy and mIoU increased to $88.67\%$ and $72.48\%$, respectively. 

\subsection{Hardware Overhead}

\begin{figure}
    \centering
    \includegraphics[width = \linewidth]{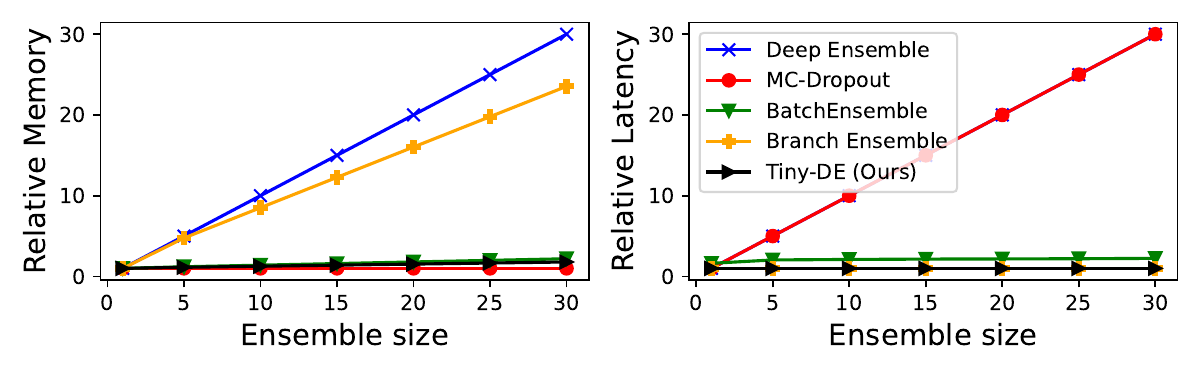}
    \caption{The inference cost in terms of memory and latency of our and related approaches w.r.t the ensemble size. The results are relative to a single model cost. 
    The testing time cost and memory cost of the naive ensemble are plotted in blue.
    }
    \label{fig:cost}
\end{figure}

Figs.~\ref{fig:cost} show the relative cost in terms of memory and latency of our approach and related approaches for the ResNet-32 topology. In terms of memory overhead, our approach has approximately the same overhead as the BatchEnsemble~\cite{batchensemble} and MC-Dropout~\cite{gal2016dropout} methods but significantly outperforms branch ensemble~\cite{rock2021efficient} and Deep Ensemble~\cite{deep_ensemble} methods. The memory overhead of the deep ensemble increases linearly with the size of the ensemble. In the branch ensemble method, the last two convolutional and final classifier layers are ensembled. Since the last two layers consume $\sim 75\%$ of the total parameters, ensembling them leads to a high memory overhead. Specifically, if batch normalization is used, our method has slightly more overhead compared to BatchEnsemble due to the requirements of running mean and variance vector storage. However, for other normalization layers that do not calculate running mean and variance, the memory overhead is the same. 

In terms of latency, our approach has the same latency as the single model, as no additional computation is required relative to the single model. Therefore, the latency is the same as the branch ensemble method. However, BatchEnsemble has additional computation requirements in the input and output of convolutional layers, leading to as much as $2\times$ latency as our method. The latency of the deep ensemble and the MC-dropout increases linearly with the size of the ensembles.

In general, our approach provides a good balance between memory and latency. In parallel mode, our approach requires \emph{one forward-passes and has approximately the same memory overhead} relative to a single model (\emph{an ideal case}). Consequently, our approach has up to $\sim M\times$ reduction in overhead.




\section{Conclusion}\label{sec:conclusion}
In this paper, we present a cost-effective ensembling method for edge AI accelerators. We introduce the Tiny-DE topology, where only normalization layers are ensembled and all ensemble members share the weights and biases. Our approach is scalable in terms of AI accelerators, datasets, NN topologies, and tasks. With an expensive evaluation, we show that our approach can estimate uncertainty effectively with up to $\sim 1\%$ improvement in accuracy, and a $17.7\%$ reduction in RMSE score on various tasks. Furthermore, our approach has up to $\sim M\times$ reduction in hardware overhead.

\bibliographystyle{IEEEtran}
\bibliography{references}

\begin{thebibliography}{10}
\providecommand{\url}[1]{#1}
\csname url@samestyle\endcsname
\providecommand{\newblock}{\relax}
\providecommand{\bibinfo}[2]{#2}
\providecommand{\BIBentrySTDinterwordspacing}{\spaceskip=0pt\relax}
\providecommand{\BIBentryALTinterwordstretchfactor}{4}
\providecommand{\BIBentryALTinterwordspacing}{\spaceskip=\fontdimen2\font plus
\BIBentryALTinterwordstretchfactor\fontdimen3\font minus \fontdimen4\font\relax}
\providecommand{\BIBforeignlanguage}[2]{{%
\expandafter\ifx\csname l@#1\endcsname\relax
\typeout{** WARNING: IEEEtran.bst: No hyphenation pattern has been}%
\typeout{** loaded for the language `#1'. Using the pattern for}%
\typeout{** the default language instead.}%
\else
\language=\csname l@#1\endcsname
\fi
#2}}
\providecommand{\BIBdecl}{\relax}
\BIBdecl

\bibitem{lecun2015deep}
Y.~LeCun, Y.~Bengio, and G.~Hinton, ``Deep learning,'' \emph{nature}, vol. 521, no. 7553, pp. 436--444, 2015.

\bibitem{carroll2011secure}
M.~Carroll, A.~Van Der~Merwe, and P.~Kotze, ``Secure cloud computing: Benefits, risks and controls,'' in \emph{2011 information security for South Africa}.\hskip 1em plus 0.5em minus 0.4em\relax IEEE, 2011, pp. 1--9.

\bibitem{loven2019edgeai}
L.~Lov{\'e}n, T.~Lepp{\"a}nen, E.~Peltonen, J.~Partala, E.~Harjula, P.~Porambage, M.~Ylianttila, and J.~Riekki, ``Edgeai: A vision for distributed, edge-native artificial intelligence in future 6g networks,'' \emph{6G Wireless Summit, March 24-26, 2019 Levi, Finland}, 2019.

\bibitem{hu2021rim}
Y.~Hu, W.~Pang, X.~Liu, R.~Ghosh, B.~Ko, W.-H. Lee, and R.~Govindan, ``Rim: Offloading inference to the edge,'' in \emph{Proceedings of the International Conference on Internet-of-Things Design and Implementation}, 2021, pp. 80--92.

\bibitem{tinyml_survey}
Y.~Abadade, A.~Temouden, H.~Bamoumen, N.~Benamar, Y.~Chtouki, and A.~S. Hafid, ``A comprehensive survey on tinyml,'' \emph{IEEE Access}, vol.~11, pp. 96\,892--96\,922, 2023.

\bibitem{hendrycks2019benchmarking}
D.~Hendrycks and T.~Dietterich, ``Benchmarking neural network robustness to common corruptions and perturbations,'' \emph{arXiv preprint arXiv:1903.12261}, 2019.

\bibitem{deep_ensemble}
B.~Lakshminarayanan, A.~Pritzel, and C.~Blundell, ``Simple and scalable predictive uncertainty estimation using deep ensembles,'' \emph{Advances in neural information processing systems}, vol.~30, 2017.

\bibitem{batchensemble}
Y.~Wen, D.~Tran, and J.~Ba, ``Batchensemble: an alternative approach to efficient ensemble and lifelong learning,'' \emph{arXiv preprint arXiv:2002.06715}, 2020.

\bibitem{abdar2021review}
M.~Abdar, F.~Pourpanah, S.~Hussain, D.~Rezazadegan, L.~Liu, M.~Ghavamzadeh, P.~Fieguth, X.~Cao, A.~Khosravi, U.~R. Acharya \emph{et~al.}, ``A review of uncertainty quantification in deep learning: Techniques, applications and challenges,'' \emph{Information fusion}, vol.~76, pp. 243--297, 2021.

\bibitem{wilson2020bayesian}
A.~G. Wilson and P.~Izmailov, ``Bayesian deep learning and a probabilistic perspective of generalization,'' \emph{Advances in neural information processing systems}, vol.~33, pp. 4697--4708, 2020.

\bibitem{gal2016dropout}
Y.~Gal and Z.~Ghahramani, ``Dropout as a bayesian approximation: Representing model uncertainty in deep learning,'' in \emph{international conference on machine learning}.\hskip 1em plus 0.5em minus 0.4em\relax PMLR, 2016, pp. 1050--1059.

\bibitem{mobiny2021dropconnect}
A.~Mobiny, P.~Yuan, S.~K. Moulik, N.~Garg, C.~C. Wu, and H.~Van~Nguyen, ``Dropconnect is effective in modeling uncertainty of bayesian deep networks,'' \emph{Scientific reports}, vol.~11, no.~1, p. 5458, 2021.

\bibitem{ahmed2023spindrop}
S.~T. Ahmed, K.~Danouchi, C.~M{\"u}nch, G.~Prenat, L.~Anghel, and M.~B. Tahoori, ``Spindrop: Dropout-based bayesian binary neural networks with spintronic implementation,'' \emph{IEEE Journal on Emerging and Selected Topics in Circuits and Systems}, vol.~13, no.~1, pp. 150--164, 2023.

\bibitem{rock2021efficient}
J.~Rock, T.~Azevedo, R.~de~Jong, D.~Ruiz-Mu{\~n}oz, and P.~Maji, ``On efficient uncertainty estimation for resource-constrained mobile applications,'' \emph{arXiv preprint arXiv:2111.09838}, 2021.

\bibitem{tuhin2022binary}
S.~Tuhin~Ahmed, K.~Danouchi, C.~M{\"u}nch, G.~Prenat, A.~Lorena, and M.~B.~Tahoori, ``Binary bayesian neural networks for efficient uncertainty estimation leveraging inherent stochasticity of spintronic devices,'' in \emph{Proceedings of the 17th ACM International Symposium on Nanoscale Architectures}, 2022, pp. 1--6.

\bibitem{ahmed2023scale}
S.~T. Ahmed, K.~Danouchi, M.~Hefenbrock, G.~Prenat, L.~Anghel, and M.~B. Tahoori, ``Scale-dropout: Estimating uncertainty in deep neural networks using stochastic scale,'' \emph{arXiv preprint arXiv:2311.15816}, 2023.

\bibitem{ahmed2023spatial}
------, ``Spatial-spindrop: Spatial dropout-based binary bayesian neural network with spintronics implementation,'' \emph{arXiv preprint arXiv:2306.10185}, 2023.

\bibitem{ioffe2015batch}
S.~Ioffe and C.~Szegedy, ``Batch normalization: Accelerating deep network training by reducing internal covariate shift,'' in \emph{International conference on machine learning}.\hskip 1em plus 0.5em minus 0.4em\relax pmlr, 2015, pp. 448--456.

\bibitem{guo2017calibration}
C.~Guo, G.~Pleiss, Y.~Sun, and K.~Q. Weinberger, ``On calibration of modern neural networks,'' in \emph{International conference on machine learning}.\hskip 1em plus 0.5em minus 0.4em\relax PMLR, 2017, pp. 1321--1330.

\bibitem{jospin2022hands}
L.~V. Jospin, H.~Laga, F.~Boussaid, W.~Buntine, and M.~Bennamoun, ``Hands-on bayesian neural networks—a tutorial for deep learning users,'' \emph{IEEE Computational Intelligence Magazine}, vol.~17, no.~2, pp. 29--48, 2022.

\bibitem{ba2016layer}
J.~L. Ba, J.~R. Kiros, and G.~E. Hinton, ``Layer normalization,'' \emph{arXiv preprint arXiv:1607.06450}, 2016.

\bibitem{ulyanov2016instance}
D.~Ulyanov, A.~Vedaldi, and V.~Lempitsky, ``Instance normalization: The missing ingredient for fast stylization,'' \emph{arXiv preprint arXiv:1607.08022}, 2016.

\bibitem{wu2018group}
Y.~Wu and K.~He, ``Group normalization,'' in \emph{Proceedings of the European conference on computer vision (ECCV)}, 2018, pp. 3--19.

\bibitem{hansen1990neural}
L.~K. Hansen and P.~Salamon, ``Neural network ensembles,'' \emph{IEEE transactions on pattern analysis and machine intelligence}, vol.~12, no.~10, pp. 993--1001, 1990.

\bibitem{dietterich2000ensemble}
T.~G. Dietterich, ``Ensemble methods in machine learning,'' in \emph{International workshop on multiple classifier systems}.\hskip 1em plus 0.5em minus 0.4em\relax Springer, 2000, pp. 1--15.

\bibitem{opitz1999popular}
D.~Opitz and R.~Maclin, ``Popular ensemble methods: An empirical study,'' \emph{Journal of artificial intelligence research}, vol.~11, pp. 169--198, 1999.

\bibitem{buciluǎ2006model}
C.~Buciluǎ, R.~Caruana, and A.~Niculescu-Mizil, ``Model compression,'' in \emph{Proceedings of the 12th ACM SIGKDD international conference on Knowledge discovery and data mining}, 2006, pp. 535--541.

\bibitem{hinton2015distilling}
G.~Hinton, O.~Vinyals, and J.~Dean, ``Distilling the knowledge in a neural network,'' \emph{arXiv preprint arXiv:1503.02531}, 2015.

\bibitem{huang2017snapshot}
G.~Huang, Y.~Li, G.~Pleiss, Z.~Liu, J.~E. Hopcroft, and K.~Q. Weinberger, ``Snapshot ensembles: Train 1, get m for free,'' \emph{arXiv preprint arXiv:1704.00109}, 2017.

\bibitem{loshchilov2016sgdr}
I.~Loshchilov and F.~Hutter, ``Sgdr: Stochastic gradient descent with warm restarts,'' \emph{arXiv preprint arXiv:1608.03983}, 2016.

\bibitem{hamdioui2015memristor}
S.~Hamdioui, L.~Xie, H.~A. Du~Nguyen, M.~Taouil, K.~Bertels, H.~Corporaal, H.~Jiao, F.~Catthoor, D.~Wouters, L.~Eike \emph{et~al.}, ``Memristor based computation-in-memory architecture for data-intensive applications,'' in \emph{2015 Design, Automation \& Test in Europe Conference \& Exhibition (DATE)}.\hskip 1em plus 0.5em minus 0.4em\relax IEEE, 2015, pp. 1718--1725.

\bibitem{yu2021compute}
S.~Yu, H.~Jiang, S.~Huang, X.~Peng, and A.~Lu, ``Compute-in-memory chips for deep learning: Recent trends and prospects,'' \emph{IEEE circuits and systems magazine}, vol.~21, no.~3, pp. 31--56, 2021.

\bibitem{mutlu2019processing}
O.~Mutlu, S.~Ghose, J.~G{\'o}mez-Luna, and R.~Ausavarungnirun, ``Processing data where it makes sense: Enabling in-memory computation,'' \emph{Microprocessors and Microsystems}, vol.~67, pp. 28--41, 2019.

\bibitem{jouppi2017datacenter}
N.~P. Jouppi, C.~Young, N.~Patil, D.~Patterson, G.~Agrawal, R.~Bajwa, S.~Bates, S.~Bhatia, N.~Boden, A.~Borchers \emph{et~al.}, ``In-datacenter performance analysis of a tensor processing unit,'' in \emph{Proceedings of the 44th annual international symposium on computer architecture}, 2017, pp. 1--12.

\bibitem{posewsky2016efficient}
T.~Posewsky and D.~Ziener, ``Efficient deep neural network acceleration through fpga-based batch processing,'' in \emph{2016 International Conference on ReConFigurable Computing and FPGAs (ReConFig)}.\hskip 1em plus 0.5em minus 0.4em\relax IEEE, 2016, pp. 1--8.

\bibitem{he2016deep}
K.~He, X.~Zhang, S.~Ren, and J.~Sun, ``Deep residual learning for image recognition,'' in \emph{Proceedings of the IEEE conference on computer vision and pattern recognition}, 2016, pp. 770--778.

\bibitem{simonyan2014very}
K.~Simonyan and A.~Zisserman, ``Very deep convolutional networks for large-scale image recognition,'' \emph{arXiv preprint arXiv:1409.1556}, 2014.

\bibitem{jha2020kvasir}
D.~Jha, P.~H. Smedsrud, M.~A. Riegler, P.~Halvorsen, T.~de~Lange, D.~Johansen, and H.~D. Johansen, ``Kvasir-seg: A segmented polyp dataset,'' in \emph{MultiMedia Modeling: 26th International Conference, MMM 2020, Daejeon, South Korea, January 5--8, 2020, Proceedings, Part II 26}.\hskip 1em plus 0.5em minus 0.4em\relax n.d.: Springer, 2020, pp. 451--462.

\bibitem{lin2017feature}
T.-Y. Lin, P.~Doll{\'a}r, R.~Girshick, K.~He, B.~Hariharan, and S.~Belongie, ``Feature pyramid networks for object detection,'' in \emph{Proceedings of the IEEE conference on computer vision and pattern recognition}, 2017, pp. 2117--2125.

\bibitem{brostow2009semantic}
G.~J. Brostow, J.~Fauqueur, and R.~Cipolla, ``Semantic object classes in video: A high-definition ground truth database,'' \emph{Pattern Recognition Letters}, 2009.

\bibitem{zhou2019unetpp}
Z.~Zhou, M.~M.~R. Siddiquee, N.~Tajbakhsh, and J.~Liang, ``Unet++: Redesigning skip connections to exploit multiscale features in image segmentation,'' \emph{IEEE transactions on medical imaging}, vol.~39, no.~6, pp. 1856--1867, 2019.

\bibitem{hernandez2015probabilistic}
J.~M. Hern{\'a}ndez-Lobato and R.~Adams, ``Probabilistic backpropagation for scalable learning of bayesian neural networks,'' in \emph{International conference on machine learning}.\hskip 1em plus 0.5em minus 0.4em\relax PMLR, 2015, pp. 1861--1869.

\end{thebibliography}

\end{document}